\documentclass{article}

\usepackage{arxiv}

\usepackage[utf8]{inputenc} 
\usepackage[T1]{fontenc}    
\usepackage{hyperref}       
\usepackage{url}            
\usepackage{booktabs}       
\usepackage{amsfonts}       
\usepackage{nicefrac}       
\usepackage{microtype}      
\usepackage{lipsum}		
\usepackage{graphicx}
\usepackage{doi}
\usepackage{xcite}

\usepackage{comment} 
\usepackage{algorithmic}
\usepackage[lined,boxed,ruled,commentsnumbered]{algorithm2e}
\usepackage{soul}
\usepackage{amsmath}
\usepackage{color}

\title{Generation of the NIR spectral Band for Satellite Images with Convolutional Neural Networks}

\author{ \href{https://orcid.org/0000-0003-2448-9907}{\includegraphics[scale=0.06]{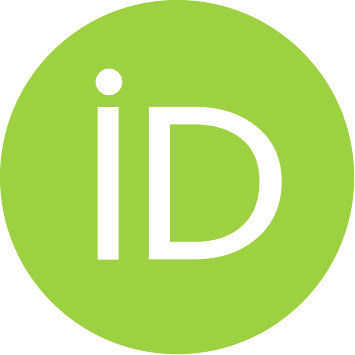}\hspace{1mm}Svetlana Illarionova} \\
	Skolkovo Institute of Science and Technology\\
	Moscow, Russia \\
	\texttt{s.illarionova@skoltech.ru} \\
	\And
	\href{https://orcid.org/0000-0003-3486-8214}{\includegraphics[scale=0.06]{orcid.pdf}\hspace{1mm}Dmitrii Shadrin} \\
	Skolkovo Institute of Science and Technology\\
	Moscow, Russia \\
	\And
	\href{https://orcid.org/0000-0003-2178-808X}{\includegraphics[scale=0.06]{orcid.pdf}\hspace{1mm}Alexey Trekin} \\
	Skolkovo Institute of Science and Technology\\
	Moscow, Russia \\
	\And
	\href{https://orcid.org/0000-0001-8565-1184}{\includegraphics[scale=0.06]{orcid.pdf}\hspace{1mm}Vladimir Ignatiev} \\
	Skolkovo Institute of Science and Technology\\
	Moscow, Russia \\
	\And
	\href{https://orcid.org/0000-0003-2071-2163}{\includegraphics[scale=0.06]{orcid.pdf}\hspace{1mm}Ivan Oseledets} \\
	Skolkovo Institute of Science and Technology\\
	Moscow, Russia \\
}

\hypersetup{
pdftitle={A template for the arxiv style},
pdfsubject={q-bio.NC, q-bio.QM},
pdfauthor={David S.~Hippocampus, Elias D.~Striatum},
pdfkeywords={First keyword, Second keyword, More},
}

\begin{document}
\maketitle

\begin{abstract}
	The near-infrared (NIR) spectral range (from 780 to 2500 nm) of the multispectral remote sensing imagery provides vital information for the landcover classification, especially concerning the vegetation assessment. Despite the usefulness of NIR, common RGB is not always accompanied by it. Modern achievements in image processing via deep neural networks allow generating artificial spectral information, such as for the image colorization problem. In this research, we aim to investigate whether this approach can produce not only visually similar images but also an artificial spectral band that can improve the performance of computer vision algorithms for solving remote sensing tasks. We study the generative adversarial network (GAN) approach in the task of the NIR band generation using just RGB channels of high-resolution satellite imagery. We evaluate the impact of a generated channel on the model performance for solving the forest segmentation task. Our results show an increase in model accuracy when using generated NIR comparing to the baseline model that uses only RGB ($0.947$ and $0.914$ F1-score accordingly). Conducted study shows the advantages of generating the extra band and its implementation in applied challenges reducing the required amount of labeled data.
\end{abstract}


\section{Introduction}
\label{sect:introduction}

Machine learning techniques allow researchers to achieve high performance in a wide range of remote sensing tasks leveraging spectral bands of different wavelengths~\cite{maxwell2018implementation}. One essential spectrum interval for the remote sensing image analysis is represented by the near-infrared (NIR) channel. 
The classical approaches in landcover classification tasks often use NIR-based spectral indices such as Normalized Difference Vegetation Index (NDVI) or Enhanced Vegetation Index (EVI) to assess the vegetation state \cite{huete1999modis}. This spectral band is widely used in many applications, including forestry \cite{li2019large}, \cite{illarionova2020neural}, agriculture \cite{kussul2017deep}, \cite{navarro2016machine}, and general landcover classification \cite{scott2017training}, \cite{2017Jiayuan}. However, there are still cases when the NIR band is not presented in the available data \cite{flood2019using}, \cite{8554605} so that the researchers can rely on RGB only. 
For example, the Maxar Open Data Program \cite{opendata} provides only RGB imagery. Many aerial imaging systems are also limited to visible wavelength ranges.   

\begin{figure*}
    \centering
    \includegraphics[width=1.\columnwidth]{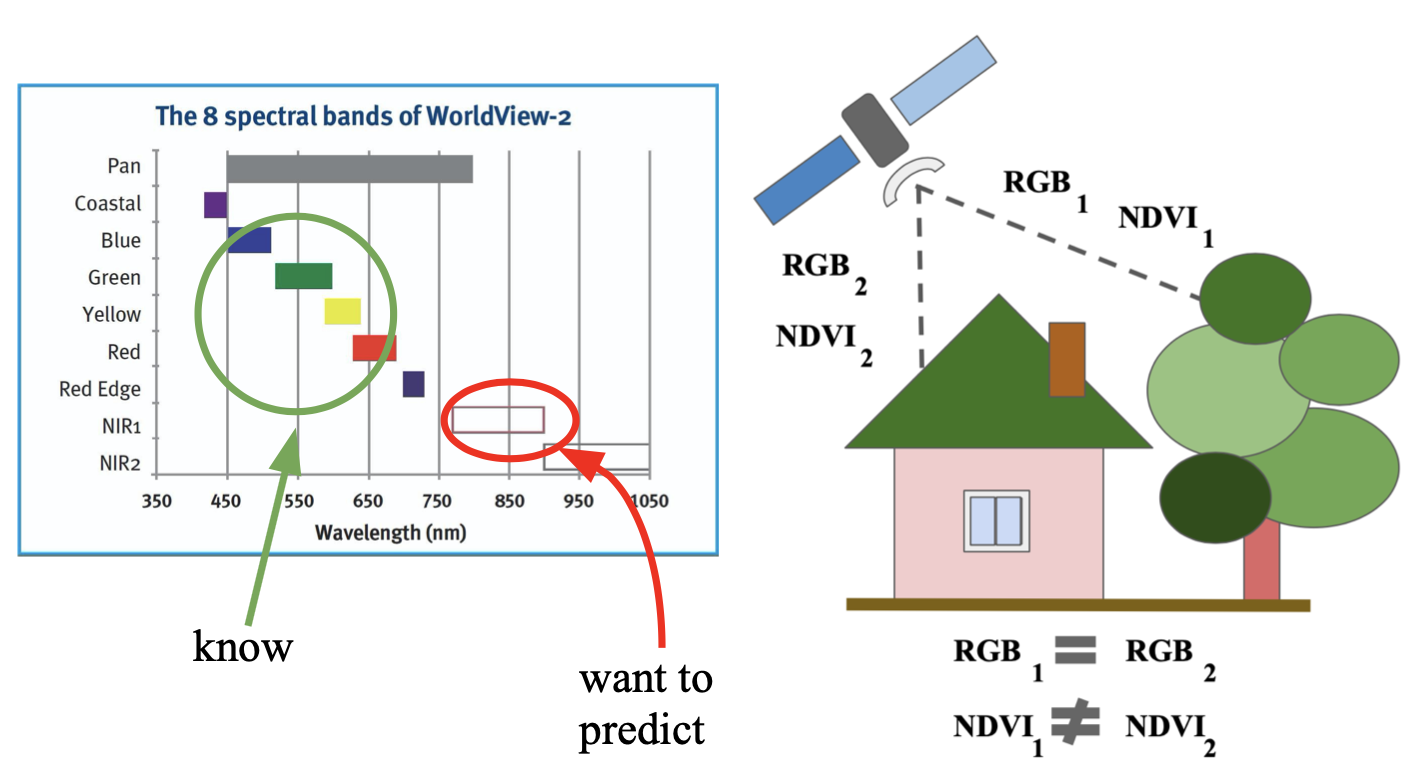}
    \caption{Objects with the same spectral values in the RGB range can belong to significantly different classes. For these objects, spectral values beyond the visible range differ. It can be illustrated using vegetation indices such as normalized difference vegetation index (NDVI) in the case with an artificial object and a plant during the vegetation period.}
    \label{fig:study_idea}
\end{figure*}

NIR band could not be extracted from RGB bands. A simple example is provided in Figure~\ref{fig:study_idea}. Both for the green tree and the green roof, RGB values can be the same. However, the value differs drastically in the NIR spectral range as the metal roof does not have vegetation properties that affect NIR. On the other hand, indirect features can be used to evaluate the NIR value. In general, all roofs have a lower NIR value than any healthy tree during the vegetation period. Therefore, it is possible to make an assumption about the NIR value based on the object's shape and texture. This study investigates how neural networks can be applied to solve the NIR generation task by learning the statistical distribution of a huge unlabeled satellite images dataset.

In \cite{de2019estimating}, a similar problem of generating the NIR channel from RGB is described. The proposed solution is based on the K-Nearest Neighbor classification algorithm and is focused on the agricultural domain. The research shows high demand for the generated NIR data, which can solve particular problems. However, the neural network approach was beyond the scope of the study both for image generation and the applied problem solution. The main differences from our problem definition are the natural structure of considered objects (fields on the images without enough high resolution do not have significant structure features, unlike vast areas covered by forest, other vegetation, and man-made regions). 

Generative adversarial networks (GAN) achieved great results in the last years~\cite{alqahtani2019applications}. The basis of this approach consists of two neural network models that are training to beat each other: the first one (generator) aims to create as realistic instances as possible, and the second one (discriminator) learns to verify whether the instance is fake or real. Conditional GANs (cGAN) have proven to  be a promising approach in various fields using additional conditions in a generation process. cGANs were implemented to solve different tasks such as image colorization \cite{nazeri2018image}, including infrared input \cite{suarez2017infrared} and remote sensing data \cite{wu2019remote, li2018multi, tang2020visualizing}, style transfer \cite{zhu2017unpaired}, \cite{isola2017image}, etc.

Pix2pix GAN described in \cite{isola2017image} proposes image-to-image translation approach. The weak side of previous works is a lack of generalization for other problems. Authors aimed to develop an efficient framework that can be successfully implemented to solve a wide variety of tasks such as image colorization, synthesizing images from a labeled map, generating land-cover maps from remote sensing images, changing the style, etc. It uses as a generator ``U-Net''-based architecture and as a discriminator convolutional ``PatchGAN''. The model was trained to estimate image originality separately for each small region. Authors used the following objective $G^* = \arg \underset{G}{\min} \text{ } \underset{D}{\max} \mathcal{L} _{cGAN} (G, D) + \lambda \mathcal{L} _{L1}(G)$ to train the model. Pix2pix approach enhancements were provided in~\cite{qu2019enhanced}.

\begin{figure}
    \centering
    \includegraphics[width=.7\columnwidth]{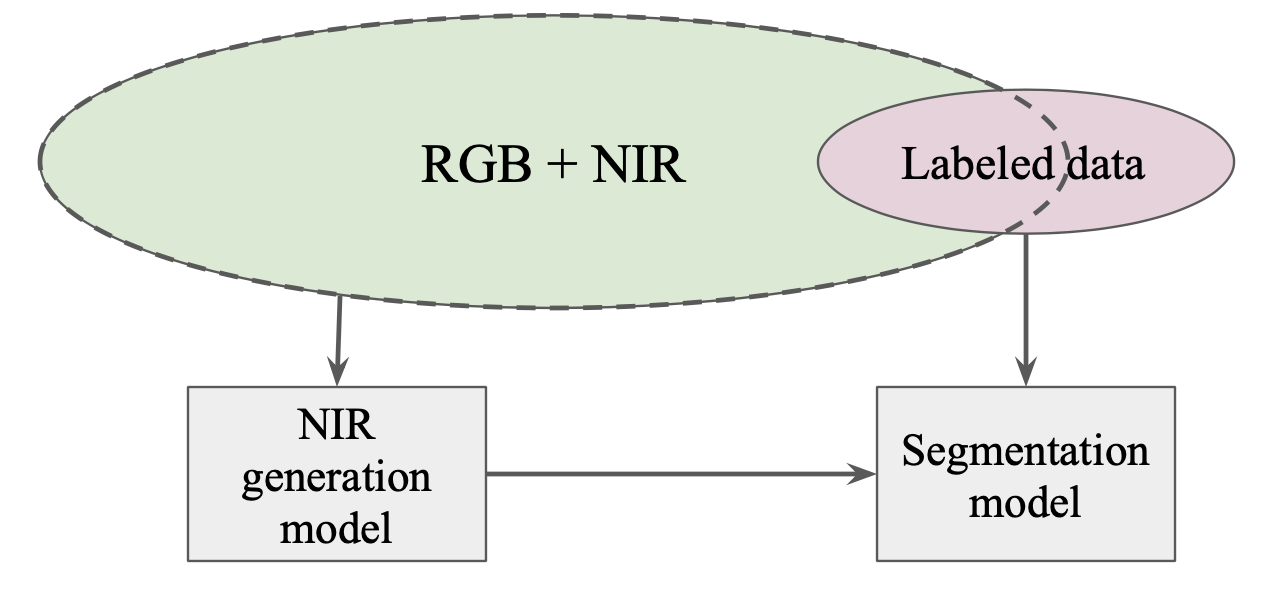}
    \caption{A huge amount of RGB + NIR data without markup that can be further leveraged in semantic segmentation tasks when NIR is not available in some particular cases.}
    \label{fig:dataset}
\end{figure}

One aforementioned prevalent computer vision task is image colorization, where it is required to obtain color images from grayscale one~\cite{wang2021overview}. In this task, cGANs take a particular condition which should be utilized for new image generation. The results for such a task can be evaluated visually. This challenge has something in common with the NIR generation problem: as an input, grayscale images are received; as an output, an RGB one is created. There is a wise versa situation with NIR: from 3 channels, we strive to get just one channel. However, unlike RGB and grayscale NIR, the main complexity does not include a mixture of RGB, and it even lies in a distant wavelength region from RGB. It makes the task more challenging. Moreover, in the colorization problem, the choice of color in some cases depends on statistical distribution in the training set (for example, the color of the car might depend on the number of cars for each color). Such mismatches in colorization might not be treated as a severe mistake, and it does not corrupt the sense of the natural source of objects or phenomena. Oppositely, there is a strong connection between chlorophyll content and the intensity of the channel value when it comes to NIR~\cite{yang2020fluorescence}. Combining RGB values with structure features such as shape and texture characteristics which the neural network can extract, we tried to generate the NIR band artificially, saved the physical sense of this channel as much as possible.            


However, in our case, we examine whether these image generation approaches can produce sufficient results for image segmentation purposes. We want to use it as a feature engineering method, generating a new feature (NIR reflectance) that is not present in the original feature space (RGB reflectances). We study original and such artificially generated NIR in the cross-domain stability problem as CNN robustness for various data is vital in the remote sensing domain~\cite{illarionova2021mixchannel}. We aim to make use of a vast amount of RGB + NIR data without markup that can be further leveraged in semantic segmentation tasks when NIR is not always available~\ref{fig:dataset}.

We aim to propose and validate an efficient approach to produce an artificial NIR band from the RGB satellite image. A state-of-the-art Pix2pix GAN technique is implemented for this task compared with a common CNN-based approach for the regression task. WorldView-2 high-resolution data is leveraged to conduct image translation from RGB to NIR with further verification on PlanetScope and Spot-5 RGB images. We also investigate how original and artificial generated NIR bands affect CNN predictions in forest segmentation tasks compared to only RGB data. The experiments involve two significant practical cases: two data sources combination (PlanetScope and Spot-5) and different amount of labeled training data (the total dataset size for the segmentation task is $500.000$ hectares). The contribution of the presented work is as follows:

\begin{itemize}
    \item The pipeline development for feature-engineering based on the NIR channel generation via cGANs;
    \item Investigation of impact of artificially generated and real NIR data on the model performance in the satellite image segmentation task. Examine the NIR channel contribution in reducing requirements for labeled dataset size with minimum quality loss and satellite cross-domain stability.
    
\end{itemize}

\section{Materials and~Methods}
\label{sect:materials_methods}  

\subsection{Dataset}
\label{sect:dataset}

\begin{figure*}
    \centering
    \includegraphics[width=1.\columnwidth]{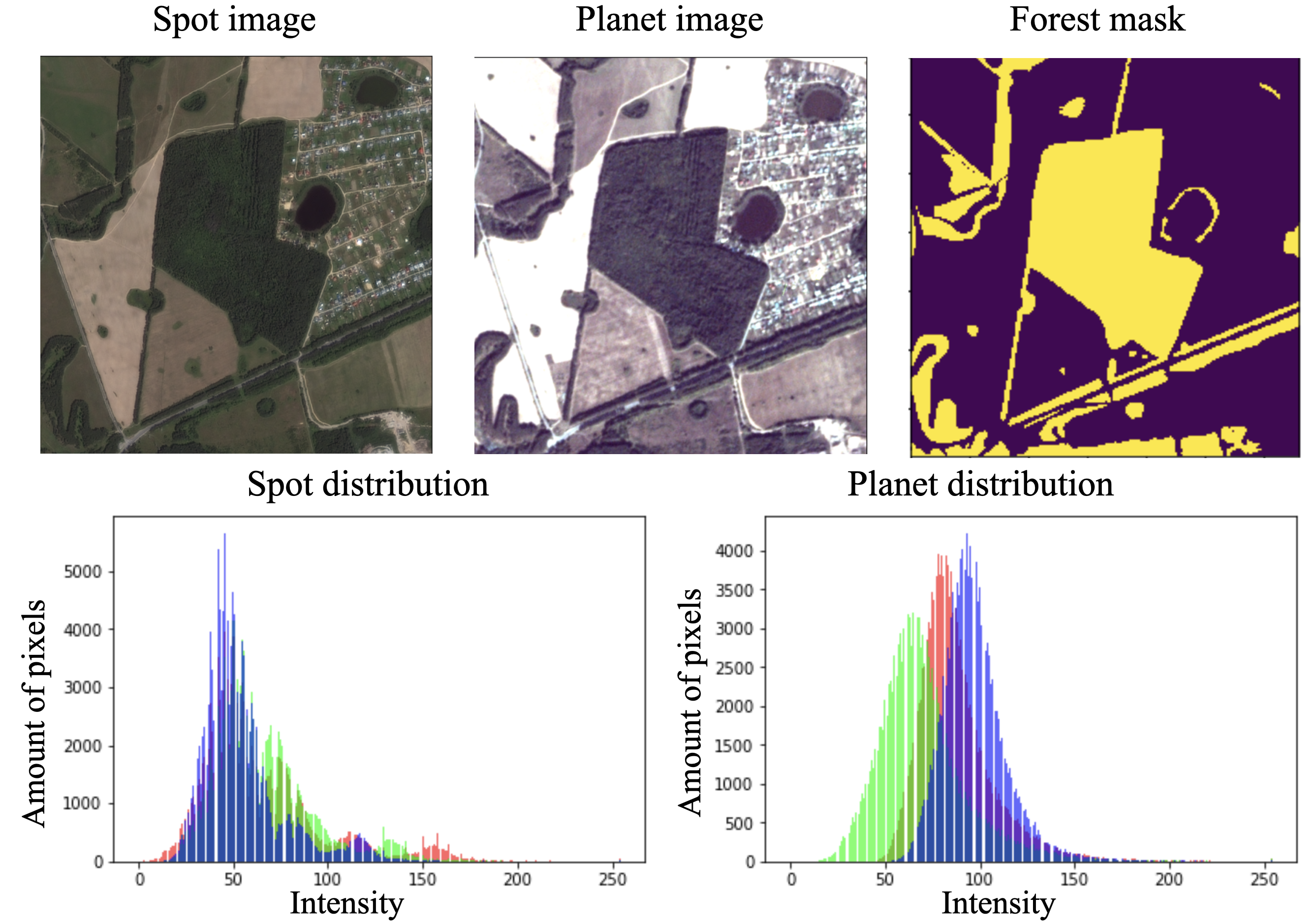}
    
    \caption{Original SPOT and Planet images (without any enhancements) and their RGB spectral values distribution. Histograms are computed within the forest area. Although the presented images are from the summer period, their spectral values differ drastically, as the histogram shows.}
    \label{fig:histogramm}
\end{figure*}

We leverage WorldView-2 satellite imagery downloaded from GBDX~\cite{gbdx} to train generative models. For experiments of forest segmentation, we use the satellite data provided by the SPOT-5~\cite{SPOT} satellite and PlanetScope~\cite{planet} satellite group. The imagery has a high spatial resolution of 2-3 meters per pixel in four spectral channels (red, green, blue, near-infrared).
Overall, two datasets are used in this work:

The first one is for a cGAN model training. It consists of RGB and NIR channels from the same satellite (WorldView-2). It covered different regions of Russia and Kazakhstan of approximately the same climate and ecological conditions. The total territory is about $900,000$ ha. It is consisted of varying land cover classes such as crops, forests, non-cultivated fields, and human-made objects. Images with dates from May to September are chosen to represent the high-vegetation period.

The second dataset is used to test the real and artificial NIR channel's influence compared to the bare RGB image. This dataset includes PlanetScope and Spot-5 imagery. The resolution of images ranges between 2 and 3 meters, depending on the view angle. The markup for the study region consists of the binary masks of the forested areas and other classes in equal proportion, covered $500,000$ ha. The labeled markup is used for the binary image segmentation problem. The region was split into test and train parts in proportions $0.25$ and $0.75$. 

\subsection{Artificial NIR channel generation}

\begin{figure*}
    \centering
    \includegraphics[width=1.\columnwidth]{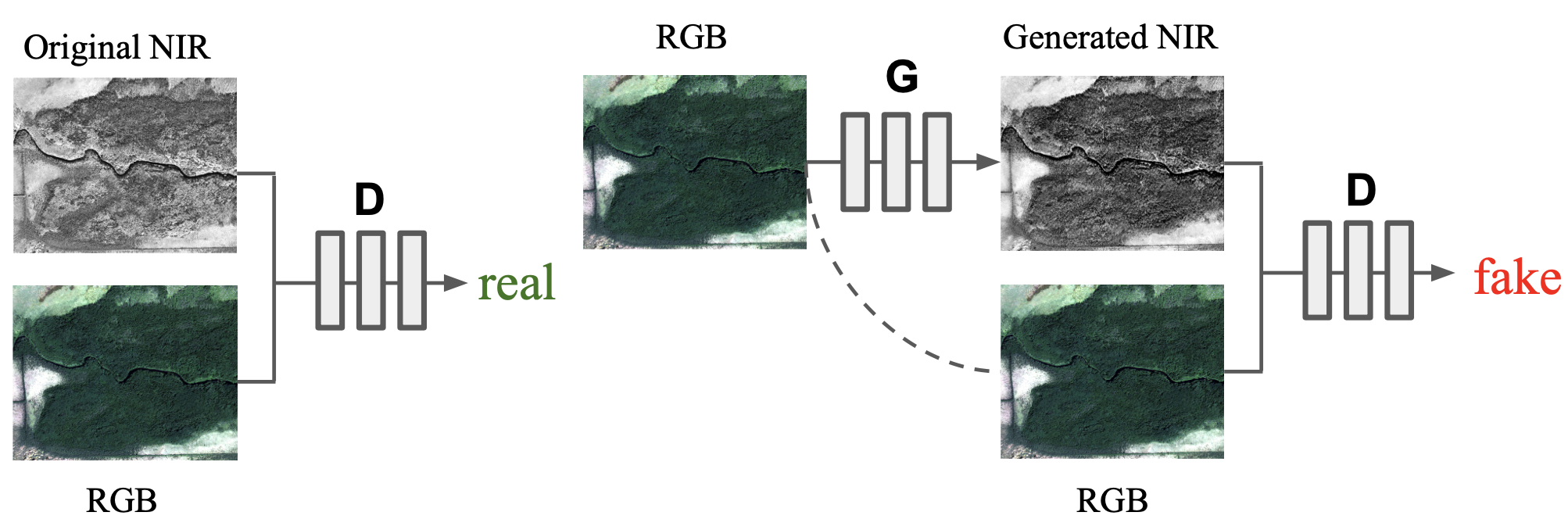}
    \caption{Training procedure for GAN using RGB image as an input and NIR band as a condition. }
    \label{fig:real}
\end{figure*}

The baseline approach aims to solve the regression problem, where the CNN is trained to predict the target band values from 3 known bands. Unlike classical machine learning techniques, which usually work just with one particular point like in \cite{de2019estimating}, CNN processes a particular neighborhood and learns how to summarize 3-dimensional information. For this task, we chose U-Net~\cite{ronneberger2015u} architecture with Resnet-34~\cite{szegedy2017inception} encoder and the linear activation function for the last layer. More details on the training process are presented in Subsection~\ref{setup}. 

In the second approach, we use conditional generative adversarial networks. We chose the pix2pix approach for this task because it performs quite well for image translation problems~\cite{salehi2020pix2pix}, \cite{ren2019two}. For the generator, we use the U-Net architecture with Resnet-34 encoder. For the discriminator, the PatchGAN described in \cite{isola2017image} with various receptive field sizes is used. The training procedure is shown in Figure \ref{fig:real}. There are two models: a generator and a discriminator. The generator is trained to create artificial NIR images, using the RGB image as a conditional input. The discriminator receives an RGB image in pair with the alleged NIR image. Then, there are few possible scenarios: NIR was original, and the discriminator succeed in ascertaining it; NIR was fake, but the discriminator failed by treating it as original; NIR was original, but the discriminator mistook for fake; NIR was fake, and the discriminator exposed it. Although this model is trained simultaneously, we strive ultimately to receive high performing generative model, as it solves the study objective. For further analysis, only the generator is considered.

The considered metrics for models performance evaluation are root mean square error (RMSE), mean absolute error (MAE), and mean bias error (MBE):

\begin{equation}
    \begin{aligned}
        RMSE = \sqrt{ \frac{\sum_{i=1}^{n}(y_i - \hat{y_i})^2}{n}}
            \end{aligned}
\end{equation}
\begin{equation}
    \begin{aligned}
        MAE = \frac{\sum_{i=1}^{n}|y_i - \hat{y_i}|}{n}
            \end{aligned}
\end{equation}
\begin{equation}
    \begin{aligned}
       MBE = \frac{\sum_{i=1}^{n}(y_i - \hat{y_i})}{n}
    \end{aligned}
\end{equation}
where $\overline{y}$ is the mean target value among all pixels, $\hat{y_i}$ is the predicted value of the $i^{th}$ pixel, $y_i$ is the target value of the $i^{th}$ pixel, and $n$  is the pixel number.

\subsection {Forest segmentation task}

To empirically evaluate the usefulness of the original and artificially generated NIR channel to real image segmentation problems, we consider the forest segmentation task with high-resolution satellite imagery. In this task, a CNN model is trained to ascribe each pixel with the forest content label.

We use the common solution for the image semantic segmentation --- U-Net~\cite{ronneberger2015u} with ResNet-34~\cite{he2016deep} encoder. The chosen architecture is widely implemented in the remote sensing domain~\cite{kattenborn2021review}. We conduct experiments with different input channels: only RGB; RGB + original NIR; RGB + generated NIR. The model output is a binary mask of forest landcover, which was evaluated against the ground truth with F1-score.

\begin{equation}
    \begin{aligned}
        precision = \frac{TP}{TP+FP}, \\
        recall = \frac{TP}{TP+FN}, \\
        F1 = \frac{2*precision*recall}{precision + recall}
    \end{aligned}
\end{equation}
where $TP$ is True Positive (number of correctly classified pixels of the given class), $FP$ is False Positive (number of pixels classified as the given class while in fact being of other class, and $FN$ is False Negative (number of pixels of the given class, missed by the method).





\subsection{NIR channel usage}
\label{sect:segmentation}

We conduct an experiment that estimates the dependency of the segmentation quality on the training dataset size in both RGB and RGB+NIR cases. We randomly split and chose $50$\% and $30$\% of the initial training dataset (test data is the same for these random splits). The same experiment is repeated both for SPOT and Planet imagery but separately for each data source. 

In the second study, we consider data from different sources (both PlanetScope and SPOT data) simultaneously. Even if we have two images of the same date, region, and resolution but from various providers, sensors systems and image preprocessing can make them radically different from each other. Images intensity distribution for Spot and Planet are shown in Figure~\ref{fig:histogramm}. Such differences can be crucial for machine vision algorithms and lead to prediction quality decrease. Therefore, it can be treated as a case of a more complex multi-domain satellite segmentation task. 
To estimate original and artificial NIR channel importance for different satellite data, we conduct the following investigation. CNN model is trained using Planet and SPOT data simultaneously. To evaluate models' performance, three test sets are considered: just Planet test images; only SPOT test images; both Planet and SPOT images. Images for Planet and Spot cover the same territory.


\subsection{Training setup}
\label{setup}
The training of all the neural network models was performed at a PC with GTX-1080Ti GPUs, using Keras~\cite{keras} with a Tensorflow~\cite{tensorflow} backend. For a simple regression model, the following training parameters were set. An optimizer RMSprop was chosen with a learning rate of $0.001$, which was reduced with patience $5$. There were $20$ epochs with $100$ steps per epoch. The batch size was specified to be $30$ with an image size of $256*256$ pixels. A model based on GAN training parameters was as follows. Loss functions were chosen binary cross-entropy and MAE. The optimizer was Adam. The batch size and image size were the same as for the simple model. The models were trained for $600$ epochs, $100$ steps per epoch, and the batch size of $30$. 

Two types of augmentation are considered in this problem: color and geometrical. For geometrical, we implement rotation, flipping, scaling, and their combinations with a probability of $0.5$. For color transformations, brightness, contrast, and MotionBlur augmentations are considered. Albumentations framework is used to perform aforementioned RGB image transformations~\cite{info11020125}. For Planet data, we also conducted a fine-tuning procedure of the pretrained generative model using a small area without the necessity of markup. For the SPOT data, there was no additional training.
\section{Results and discussion}

\begin{figure*}
    \begin{center}
        
    \includegraphics[width=1.\columnwidth]{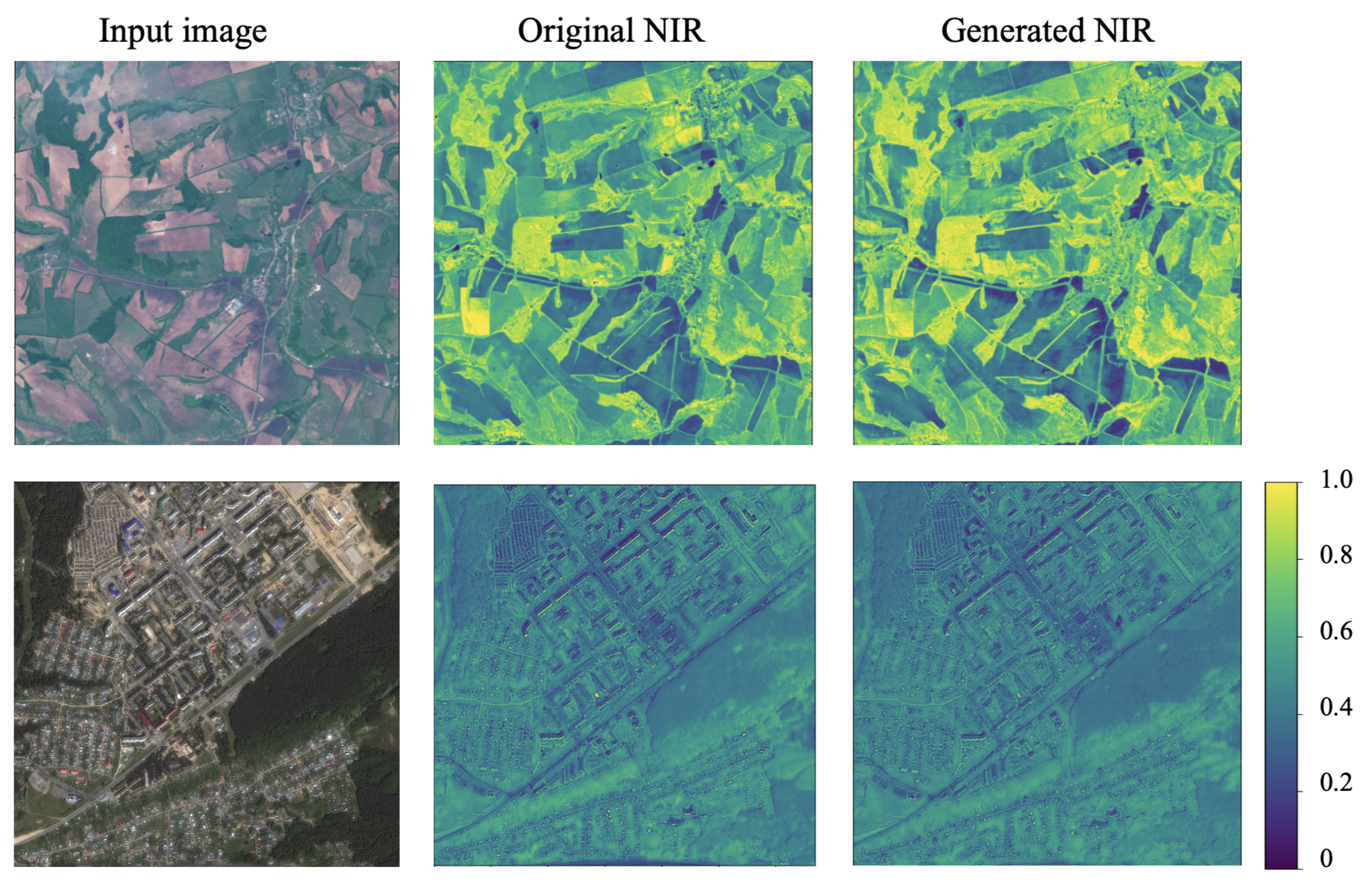}
    
    \end{center}
    \caption{Example of generated NIR on the test set. The first row presents the SPOT image; the second row is the WorldView image.}
    \label{fig:generated_nir}
\end{figure*}

\begin{figure*}
    \centering
    \includegraphics[width=1.\columnwidth]{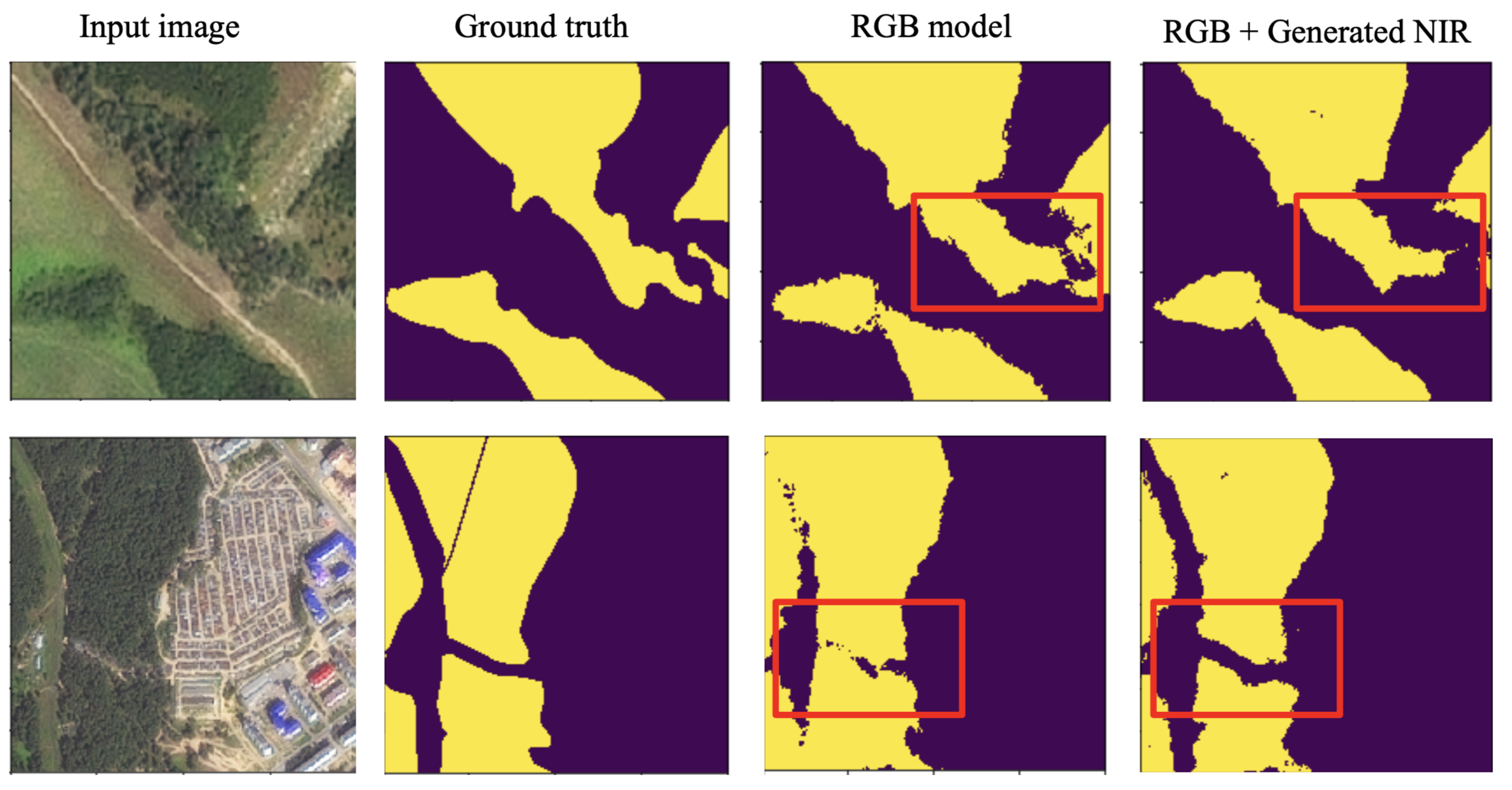}

     \caption{Forest segmentation predictions on the test regions (SPOT). One model was trained just on RGB images; another model used RGB + generated NIR.}
    \label{fig:segm_spot}
\end{figure*}

\begin{table}[h!]
\caption{Error of artificial NIR band for test WorldView, SPOT and Planet imagery}
    \label{tab:spot_planet_pr}
\begin{center}
\begin{tabular}{l|c|c|r}
\hline
    & MAE & RMSE  & Mean Bias \\

\hline
WorldView & 0.09 & 0.31& 0.058 \\
\hline
SPOT & 0.037  & 0.194& -0.0029 \\
Planet & 0.16 & 0.41 & 0.088 \\
\hline
\end{tabular}
\end{center}
\end{table}

\begin{table}[h!]
\caption{Results for forest segmentation experiments with different dataset sizes. F1-score for SPOT and Planet on the test set. The entire data size is $500,000$ ha.}
    \label{tab:different_size}
    \begin{center}
\begin{tabular}{l|c|c|c|r}
\hline
   & bands & all data & $1/2$ & $1/3$ \\
\hline
SPOT &RGB       & 0.97 & 0.956 & 0.942    \\
&RGB + NIR & 0.97 & 0.963 & 0.961     \\
\hline
Planet &RGB       & 0.939 & 0.933 & 0.874     \\
&RGB + NIR & 0.95  & 0.942 & 0.927     \\
\hline
\end{tabular}
\end{center}
\end{table}



Results for NIR generation by cGAN are presented in Table~\ref{tab:spot_planet_pr} for WorldView, SPOT, and Planet satellite data. All values for real and generated NIR are in the range $[0, 1]$. The simple CNN regression approach shows significantly poor results (MAE is $0.21$ for WorldView). Therefore, this approach is not involved in the future analysis in this study. The principal difference between cGANs and the regression CNN model is the type of loss function. As our experiments show, both MAE and MSE loss in the regression CNN model lead to local optimum, far from the global one. It can be affected by RGB values distribution. cGAN allows getting significantly closer to the real NIR values. 

Another approach to evaluate the generated NIR band involves the forest segmentation task. The segmentation model is trained on the original NIR channels to predict the forest segmentation mask using RGB + generated NIR. Results are presented in Table \ref{tab:result}, where we can see that the additional NIR channel improves the cross-domain stability of the model. The example of segmentation prediction is shown in Figure~\ref{fig:segm_spot}. Model using the generated NIR provides more accurate results than a model trained only on RGB bands. The original NIR usage leads to $0.953$ F1-score; the generated NIR leads to $0.947$ F1-score compared with $0.914$ score for the model using only RGB bands. The predicted NIR channel is shown in Figure \ref{fig:generated_nir}, which confirms a high level of similarity between generated and original bands.
Therefore, this approach allows more efficient CNN model usage in practical cases when data from different Basemaps are processed, and cross-domain tasks occur.

Results for different dataset sizes are presented in Table \ref{tab:different_size} and show that leveraging the NIR channel is beneficial in the case of smaller dataset sizes, whereas its effect decreases with the growing amount of the training data.

\begin{table}[h!]
\begin{center}
\caption{Results for forest segmentation experiments with different data sources. Both the RGB model and RGB + NIR model are trained on Planet and Spot images simultaneously. F1-score is computed on the test set individually for Planet and Spot and for the joined Planet and Spot test set.}
    \label{tab:result}
\begin{tabular}{l|l|l|l}
\hline
 Test images   & RGB & RGB + NIR & RGB + artificial NIR  \\

\hline
SPOT   & 0.954 & 0.961 &  0.96 \\
Planet & 0.857 & 0.939 &  0.936\\
SPOT + Planet   & 0.932 & 0.96 & 0.945 \\ 
\hline
Average & 0.914 & 0.953 &  0.947 \\
 & & \textcolor{blue}{(+0.039)} &  \textcolor{blue}{(+0.033)} \\
\end{tabular}

\end{center}
\end{table}

Experiments indicate that generated NIR gives additional information to the segmentation model. We assume that the generative model incorporates the hidden statistical connections between the spectral channels that can be learned from the significant amount of the real RGB+NIR data. As opposed to the segmentation or classification approach, the channel generation does not require the manual ground truth markup so that the dataset size can be significantly increased. Therefore, this approach can be used as a feature engineering tool that creates a new feature similar to the NIR band of multispectral remote sensing imagery.

The study's possible direction is to implement this feature engineering approach to other remote sensing tasks such as agriculture classification and land-cover semantic segmentation. Also, it seems to be promising to improve challenges when only drones' RGB channels are available. Another direction is to combine this feature engineering approach with different augmentation techniques for remote sensing tasks~\cite{yu2017deep}, \cite{illarionova2021object}.

It is promising to investigate the application of NIR generation methods beyond remote sensing problems in future works. Since NIR provides valuable auxiliary data in plant phenotyping tasks, NIR generation can be extended for greenhouses where high precision is vital~\cite{Nesteruk2021Image_Compression}.

\section{Conclusion}
\label{sect:conclusion}
NIR band contains essential properties for landcover tasks. However, in particular cases, this band is not available. This study investigates pix2pix cGAN implementation for image-to-image translation from RGB space imagery to the NIR band. We propose an efficient feature engineering approach based on an artificial NIR band generation.  We conduct forest segmentation experiments to assess the NIR band importance in cases of small datasets and different satellite data sources. The proposed approach improves the model robustness to data source diversity and reduces the requirement to marked dataset size, which is crucial for machine learning challenges. We assume that this data generation strategy can be implemented in practical tasks that require the NIR channel. This method can be extended to other spectral channels and remote sensing data sources.

\vspace{6pt}

\bibliographystyle{abbrv}
\bibliography{template}
\end{document}